\documentclass[journal]{IEEEtran}
  \usepackage{cite}
  \usepackage{hyperref}

\ifCLASSINFOpdf
  \usepackage[pdftex]{graphicx}
  \usepackage{subcaption}
  \usepackage{xcolor}
  \usepackage{booktabs}
  \usepackage{multirow}
  \usepackage{tabularx}
  \usepackage{adjustbox}
  \usepackage{afterpage}
  \usepackage{colortbl}  
  \usepackage{color}
  \usepackage{tablefootnote}
  \usepackage{caption}
  \usepackage{threeparttable}
  \usepackage{ragged2e}
  \usepackage{makecell}
  \definecolor{deepred}{rgb}{0,0,0}
  \definecolor{deepred2}{rgb}{0,0,0}
\else
\fi
\usepackage{amsfonts, amssymb, amsmath, bm}
\begin{document}
%
\title{\textcolor{deepred}{A Lightweight Sparse Focus Transformer for Remote Sensing Image Change Captioning}}
%
%
%


\author{Dongwei Sun, Yajie Bao, Junmin Liu, Xiangyong Cao
\thanks{Dongwei Sun, Yajie Bao and Xiangyong Cao are with the School of Computer Science and Technology and the Ministry of Education Key Lab for Intelligent Networks and Network Security, Xi’an Jiaotong University, Xi’an 710049, P.R. China (Email:sundongwei@outlook.com, byjhhhh0409@gmail.com, caoxiangyong@xjtu.edu.cn) (\emph{Corresponding author: Xiangyong Cao.}).}
\thanks{Junmin Liu is with Department of Information Science
School of Mathematics and Statistics, Xi’an Jiaotong University
No.28, Xianning West Road, Xi’an, Shaanxi, 710049, P.R. China(Email:junminliu@mail.xjtu.edu.cn)}

}

%
%

\markboth{Journal of \LaTeX\ Class Files,~Vol.~13, No.~9, February~2024}%
{Shell \MakeLowercase{\textit{et al.}}: Bare Demo of IEEEtran.cls for Journals}
%



\maketitle

\begin{abstract}
Remote sensing image change captioning (RSICC) aims to automatically generate sentences that describe content differences in remote sensing bitemporal images. Recently, attention-based transformers have become a prevalent idea for capturing the features of global change. However, existing transformer-based RSICC methods face challenges, e.g., high parameters and high computational complexity caused by the self-attention operation in the transformer encoder component. To alleviate these issues, this paper proposes a Sparse Focus Transformer (SFT) for the RSICC task. Specifically, the SFT network consists of three main components, i.e. a high-level features extractor based on a convolutional neural network (CNN), a sparse focus attention mechanism-based transformer encoder network designed to locate and capture changing regions in dual-temporal images, and a description decoder that embeds images and words to generate sentences for captioning differences. The proposed SFT network can reduce the parameter number and computational complexity by incorporating a sparse attention mechanism within the transformer encoder network. Experimental results on various datasets demonstrate that even with a reduction of over 90\% in parameters and computational complexity for the transformer encoder, our proposed network can still obtain competitive performance compared to other state-of-the-art RSICC methods. The code is available at \href{https://github.com/sundongwei/SFT_chag2cap}{Lite\_Chag2cap}.
\end{abstract}


%
\IEEEpeerreviewmaketitle

\section{Introduction}
\begin{figure}[htp]
    \centering
    \includegraphics[width=\linewidth]{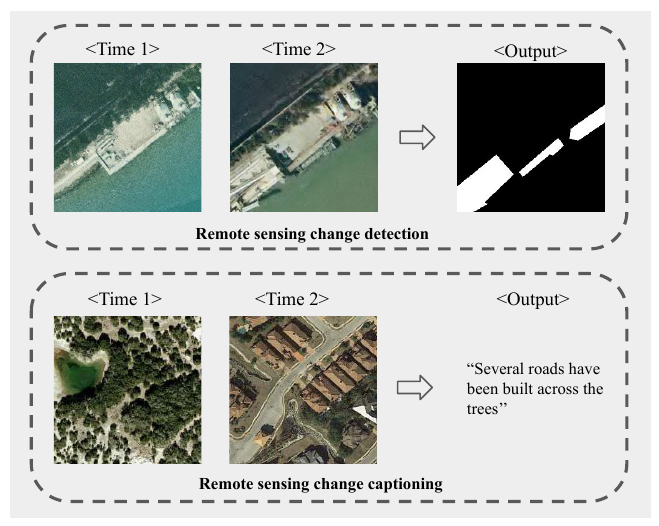}
    \caption{\textcolor{deepred}{Comparison between change detection and change captioning for remote sensing images. The former (top of the figure) represents the detected change areas in image form, while the latter (bottom of the figure) expresses changes in remote sensing images through human-readable language.}}
    \label{fig0}
\end{figure}

\IEEEPARstart{R}{emote} sensing image change captioning (RSICC) has been a hot research topic~\cite{IEEEhowto:Donahue,IEEEhowto:Lu,IEEEhowto:xu,Anderson}. This emerging research area seeks to provide meaningful descriptions of alterations within scenes, which is a valuable tool for understanding changes in land cover over time. The dynamic nature of remote sensing data, with its multi-temporal characteristics, presents unique challenges and opportunities for change captioning, different from the traditional change detection task as shown in Fig.~\ref{fig0}. Specifically, the output of change captioning is text descriptions, while the output of change detection is the image-changed region. The advent of multitemporal remote sensing data availability has sparked a growing interest in using change captioning to study changes in land cover~\cite{park,graves,park2}. In recent years, a series of RSICC methods have been proposed, and these methods can be categorized into three types: the first type primarily relies on traditional machine learning algorithms, such as Support Vector Machines(SVM), to generate corresponding descriptions; 
the second type predominantly employs traditional convolutional neural networks (CNNs) or recurrent neural networks (RNNs) methods for generating change descriptions; the third type encompasses transformer network methods based on attention mechanisms.


The first two categories of methods\cite{hoxha}\cite{chouaf} based on SVM and RNNs perform poorly in terms of captioning accuracy and precision, rendering them of limited practical value. Conversely, the third category\cite{liuc} and \cite{changs} based on attention mechanisms has significantly enhanced accuracy in change description, demonstrating suitability for practical applications. Nonetheless, due to the high complexity and parameter count of attention mechanisms, particularly in Transformer architectures, deployment and practical application in industrial settings with limited computational resources remain challenging. Therefore, against this backdrop, there is an urgent need to devise a lightweight algorithm incorporating attention mechanisms, ensuring both high accuracy and practical deployability.
\begin{figure}[ht]
\centering
\includegraphics[width=3.0in]{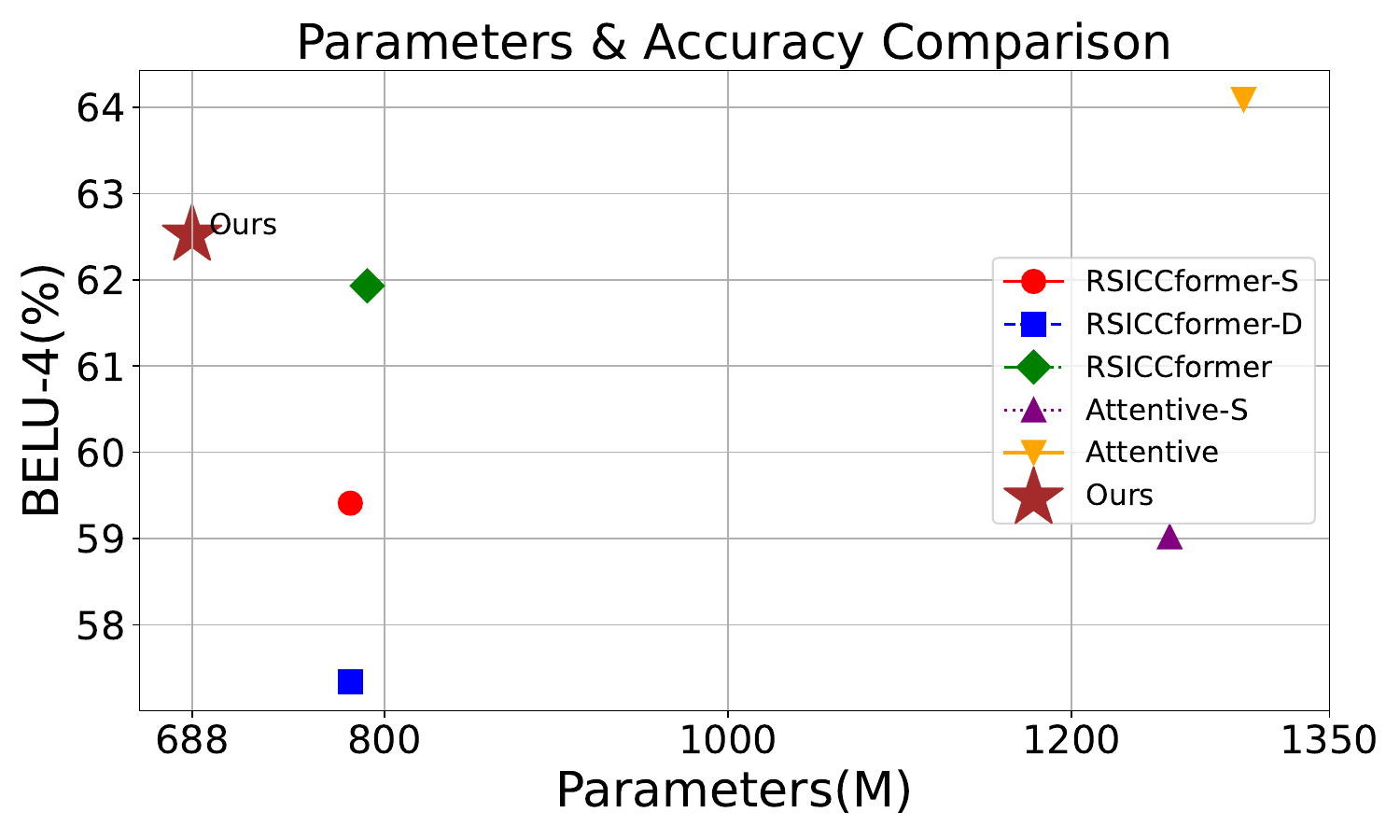}
\caption{Illustration of Algorithmic Evaluation: Computational Efficiency (Parameter Count) and Predictive Accuracy.}
\label{fig1}
\end{figure}
To address the aforementioned issues, this paper proposes a Sparse Focus Transformer (SFT) network for remote sensing change captioning. As illustrated in Fig.~\ref{fig1}, our proposed SFT can not only ensure high-precision output results, characterized by accurate descriptive text, but also significantly reduce the parameter count compared to previous approaches.

Our method is primarily inspired by the work of \cite{ily}, which explores sparse factorizations of the attention matrix, reducing computational complexity from $\mathcal{O}(n^2) $ to $\mathcal{O}(n\sqrt{n}) $ in order to generate long sequences with lower computation in natural language processing (NLP).
After further refinement and enhancement of this concept, we have introduced it into change captioning, achieving the first instance of model compression in the multimodal domain of remote sensing image change detection captioning, Our work primarily contributes in the following three aspects:
\begin{enumerate}
    \item[$\cdot$]We adapt the sparse factorizations of the attention matrix approach from generating long sequential text to the task of remote sensing image change detection, aiming to establish a sparse attention mechanism for locating change regions.
    \item[$\cdot$]We construct a sparse focus transformer tailored for the task of change captioning in remote sensing images, significantly eliminating redundancy in multimodal models, thereby achieving better modality representation and subsequent fusion.
    \item[$\cdot$]Extensive validation on datasets has been conducted, resulting not only in high-accuracy outputs but also in the lowest computational complexity and parameter count among current methodologies in this field.
\end{enumerate}

The subsequent sections of the paper are structured as follows. Section \ref{sec2} offers an overview of the related work. The proposed method is expounded upon in Section \ref{sec3}. Section \ref{sec4} delves into the experiments, providing detailed descriptions and discussing the results obtained. Lastly, a concise conclusion is drawn in Section \ref{sec5}.
\section{Related work}
\label{sec2}
\subsection{Efficient Attention Architecture}
The attention mechanism\cite{vas}, along with the transformer architecture\cite{dos} it underpins, has emerged as a pivotal advancement in the field of deep learning in recent years. Operating within the encoder-decoder paradigm, it has found widespread application across various tasks. In the subsequent, we will explore scholarly articles dedicated to enhancing the efficacy of attention mechanisms and Vision Transformers (ViTs).

\textbf{Enhancing Locality.} Sparse Transformer's\cite{ily} principle involves decomposing the computation of full attention into several faster attention operations, which, when combined, approximate dense attention computation. LongFormer \cite{bel}, it introduces a spatio-temporal complexity linearly dependent on the length of the text sequence in Self-Attention, aimed at ensuring the model can effectively utilize lower spatio-temporal complexity for modeling lengthy documents. LeViT \cite{grah} and MobileViT \cite{mehta} adopted hybrid architectures featuring stacked convolution layers, effectively diminishing the number of features during the initial layer. Twins \cite{chux} employed a strategy of alternating between local and global attention layers to enhance performance. RegionViT \cite{chenc} introduced the concept of regional tokens and local tokens, thereby enriching local context with global information. Huang \cite{huangzl} and \cite{hoj} proposed axial Transformers, a self-attention-based  model for images process and other data organized as high dimensional tensors. CrossViT \cite{cf} utilizes distinct processing for small-patch and large-patch tokens, integrating them via multiple attention mechanisms. Pan et al. \cite{panz} introduced the HiLo attention method to segregate high and low-frequency patterns within an attention layer, dividing the heads into two groups, each equipped with specialized operations tailored to local window focus.

\textbf{Faster Attention.} Swin \cite{liuz} and Cswin \cite{dongx} incorporated local attention within a window and introduced a shifted window partitioning method to enable cross-window connections. Shuffle Transformer \cite{huangz} and Msg-transformer \cite{fangj} employed spatial shuffle operations as alternatives to shifted window partitioning, facilitating cross-window connections.FasterViT \cite{hatam} introduced hierarchical attention, breaking down global self-attention into multi-level attention components. FLatten Transformer \cite{dhan} integrated depth-wise convolution in conjunction with linear attention mechanisms to address the challenge of maintaining diversity in output features across different positions.

A summary of the efficient attention work reveals that our sparse focus attention not only enhances local information in images but also enables the accumulation of long-distance context. Simultaneously, it is implemented with simplicity and efficiency in both its realization and speed.
\subsection{Remote Sensing Image Change Captioning}
Hoxha et al. \cite{hoxha} proposed early and late feature fusion strategies to integrate bitemporal visual features, employing an RNN and a multi-class Support Vector Machine (SVM) decoder to generate change captions. Chouaf et al. \cite{chouaf} pioneered the Remote Sensing Image Change Captioning(RSICC) task by employing a CNN as a visual encoder to capture temporal scene changes, while adopting an RNN as a decoder to generate change descriptions. Liu et al. \cite{liuc} recently introduced a Transformer-based encoder-decoder framework for the RSICC task. Their approach involves utilizing a dual-branch Transformer encoder to detect scene changes and proposing a multistage fusion module to merge multi-layer features for change description generation. Furthermore, Liu et al. \cite{liuc2} enhanced their method by incorporating progressive difference perception Transformer layers to capture high-level and low-level semantic change information. Additionally, Liu et al. \cite{liuc3} proposed a prompt-based approach leveraging pre-trained large language models (LLMs) for RSICC tasks. They employed visual features, change classes, and language representations as input prompts to a frozen LLM for the generation of change captions. Chang \cite{changs} proposed an attentive network for remote sensing change captioning, called Chg2Cap, which utilizes the power of transformer models in natural language processing.

Through an investigation of the aforementioned RSICC efforts, it was observed that prior endeavors have attempted various methods to address the task. However, they \textcolor{deepred2}{did not consider} practical industrial applications; either the accuracy was insufficient or the complexity too high. Consequently, we propose a lightweight algorithm that not only achieves higher accuracy but also meets the requirements for real-world application scenarios.
\section{Methodology}
\label{sec3}
In this section, we will introduce our proposed lightweight transformer model for the RSICC task. 

\subsection{Overview}
\begin{figure*}[ht]
    \centering
    \includegraphics[width=\textwidth]{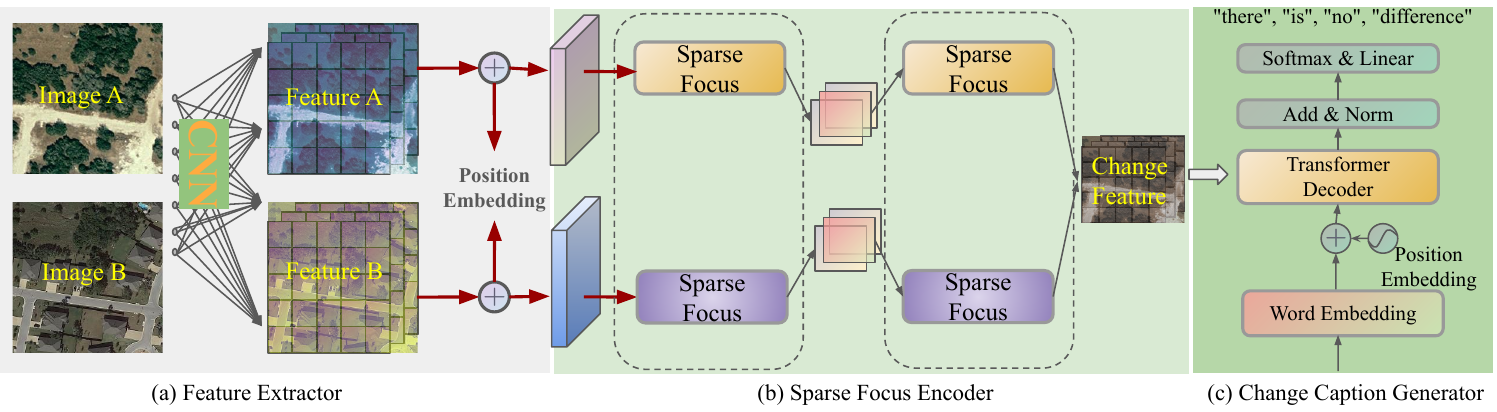}
    \caption{\textcolor{deepred}{The overall framework of the proposed Sparse Focus Transformer method comprises three components: (a) Feature Extractor: a CNN-based, weight-shared feature extractor, primarily utilizing ResNet101 in this study to extract coarse change features from bitemporal remote sensing images; (b) Sparse Focus Encoder: an encoder designed to finely capture and localize change features in remote sensing images, based on the proposed sparse focus attention mechanism; (c) Change Caption Generator: a decoder designed to generate the final change captioning for remote sensing images by accepting both the change feature embeddings and word embeddings.}}
    \label{fig2}
\end{figure*}

First, we will provide an overview of our model. As shown in Fig.~\ref{fig2}, our proposed model contains two phases, i.e., training phase and testing phase. In the following, we will introduce the two phases seperately.

\textbf{Training phase.} The Sparse Focus Transformer, as illustrated in Figure \ref{fig2}, consists primarily of three components: a CNN feature extractor, \textcolor{deepred}{where we employ ResNet101 as extractor,} responsible for extracting generic representation information from the bitemporal images; an image encoder based on sparse focus attention, designed to extract features of the changed regions using the sparse attention mechanism proposed in this paper; and finally, the caption decoder utilizes both image embeddings and word embeddings to generate predicted change captions, capturing the inter-relationship between them. The training phase primarily follows the following procedure: First, the images $\mathrm{I_1}$ and $\mathrm{I_2}$ at time instances $\mathrm{Time 1}$ and $\mathrm{Time 2}$, along with their manually annotated corresponding change descriptions $\bm{Text=(t_1,t_2,...,t_n)}$, are taken as input, where $n$ is the length of the sequence. Subsequently, a CNN, employing weight sharing, serves as the extractor for image representation, and the output is fed into the subsequent Sparse Focus Transformer (SFT) phase for the localization and capture of changed regions. Finally, this change feature, along with the corresponding $\bm{Text}$ after embedding, is passed to the caption decoder for similarity calculation and training.

\textbf{Test phase.}The predicted captions are generated through an autoregressive process solely based on the input image pairs. Autoregression in the context of image captioning entails generating a caption incrementally, with each word predicted based on the preceding words. To elaborate, the caption generation process commences with the initialization of the ``START" token, and subsequent predictions are contingent upon the previously generated words.

\subsection{Theory of Sparse Attention}
\textbf{Analysis of Complexity.} The transformer architecture primarily consists of modules such as feedforward neural networks, activation functions, self-attention mechanisms, and others. Among these, the primary attention module is described by the following formulation: \\
\begin{equation}
    Attention(\bm{Q, K, V}) = Softmax(\frac{\bm{QK}^T}{\sqrt{d}})\bm{V}
\end{equation}
where $\bm{Q, K, V}$ are query, key and value matrix, respectively. $\mathrm{d}$ is the feature’s dimension. In this computational paradigm, the most computationally intensive aspect is focused on the operation involving the matrix $\bm{QK}^T$, reaching a complexity of $\mathcal{O}(n^2)$. The core principle of Sparse Transformers is to involve only the pixels that influence the current pixel in the calculation of self-attention.

\subsection{Sparse Focus Network}
\begin{figure}[htb]
\centering
\subcaptionbox{conventional attention kernel}{\includegraphics[width=0.3\linewidth]{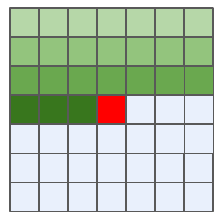}}
\hfill
\subcaptionbox{row attention kernel}{\includegraphics[width=0.3\linewidth]{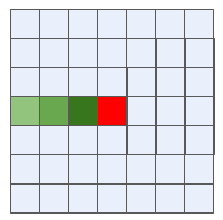}}
\hfill
\subcaptionbox{column attention kernel}{\includegraphics[width=0.3\linewidth]{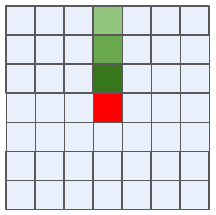}}
\caption{visualization of attention kernel.}
\label{fig4}
\end{figure}

Through the theoretical analysis, we visualize the sparse factorizations of the attention kernels. Figure \ref{fig4}(a) illustrates the computation of a conventional attention kernel, where the prediction of the current pixel (depicted in deep red) involves the utilization of pixels within a certain range (light-colored region). Figure \ref{fig4}(b) demonstrates a row-wise attention kernel, where the prediction of the current pixel (depicted in deep red) exclusively considers pixels within the same row (light-colored region). Similarly, Figure \ref{fig4}(c) follows the same rationale, illustrating a column-wise attention kernel where the prediction of the current pixel only involves pixels within the same column.

\begin{figure}[htb]
	\begin{minipage}[t]{\linewidth}
		\centering
		\includegraphics[width=\linewidth]{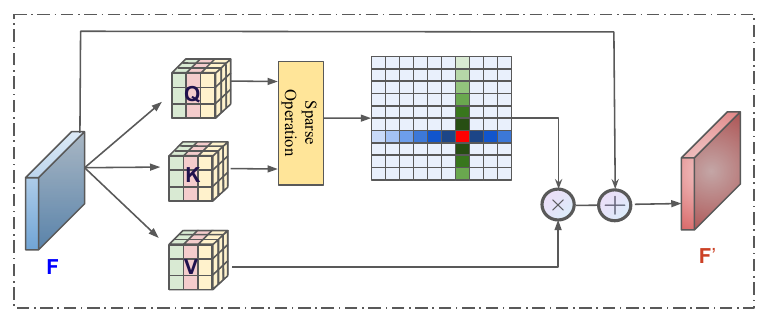}
		\caption{Sparse Focus full attention, refers to the scenario where both the row-wise attention and column-wise attention lengths for each point are equal to the entire length of the feature map. }
		\label{fig5}
	\end{minipage}
	\begin{minipage}[t]{\linewidth}
		\centering
		\includegraphics[width=\linewidth]{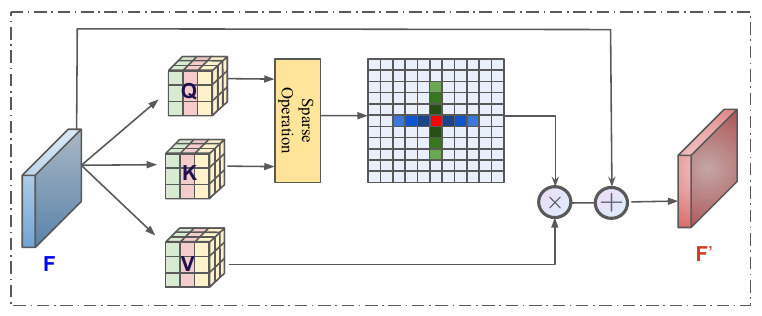}
		\caption{Sparse Focus fixed attention, refers to the scenario where both the row-wise attention and column-wise attention lengths for each point are fixed  length.}
		\label{fig6}
	\end{minipage}
\end{figure}

Based on the analysis of attention kernels presented above, we have developed Sparse Focus Attention. This module integrates both horizontal and vertical attention kernels into a novel attention kernel. In Sparse Focus Attention, we have devised two distinct types of attention kernels, as depicted in Figure \ref{fig5} and Figure \ref{fig6}. These are distinguished based on how they compute the length of rows and columns influencing the prediction of pixels. They are termed ``full length" and ``fixed length" respectively. 
The former calculates the lengths of rows and columns affecting the current pixel $P$ across the entire $W \times H$ feature map, whereas the latter computes a fixed length of pixel attention within the entire feature map.

In the ``full length" approach, the initial phase involves processing the feature map $\bm{F} \in \mathbb{R}^{C \times W \times H}$, which has been extracted by the feature extractor. This feature map $\bm{F}$ is first subjected to convolutional neural network operations, producing three feature matrices: $\bm{Q}$, $\bm{K}$, and $\bm{V}$. Specifically, $\bm{Q}$ and $\bm{K} \in \mathbb{R}^{C' \times W \times H}$, where $C'$ represents a reduced channel dimension compared to $C$.

Utilizing the sparse focus method on $\bm{Q}$ and $\bm{K}$, which involves overlaying the calculation methods of horizontal and vertical attention kernels, an attention map $\mathrm{\mathbf{A}} \in \mathbb{R}^{(H+W-1) \times (W\times H)}$ is obtained based on $\bm{Q}$ and $\bm{K}$. Through the entire channel dimension we \textcolor{deepred}{could} obtain
a vector $\mathrm{\mathbf{Q}}_p \in \mathbb{R}^{C'}$ at each pixel $\mathrm{p}$ on the matrix $\bm{Q}$. For matrix $\bm{K}$, we can obtain a vector $\mathrm{\mathbf{K}}_p \in \mathbb{R}^{(H+W-1) \times (C')}$ composed of attention kernels with the same rows and columns as matrix $\bm{Q}$. Then, the attention map $\mathrm{\mathbf{A}}$ is calculated as 
\begin{equation}
    \mathrm{\mathbf{A}} = \mathrm{Softmax}( \mathrm{\mathbf{Q}}_p \cdot \mathrm{\mathbf{K}}_{i,p}^{T} )
\end{equation}
where $\mathbf{K}_{i,p}$ is the i-th element of $\mathbf{K}_p$, applying softmax operation on it, we obtain the final attention map $\mathrm{\mathbf{A}} \in \mathbb{R}^{(H+W-1) \times (W\times H)}$. Next, we apply the conv layer on feature $\bm{F}$ to generate matrix $\mathrm{V} \in \mathbb{R}^{C\times W\times H}$. The vector $\mathrm{\mathbf{V}}_p$
is a collection of feature vectors in $\bm{V}$ which are in the same row or column corresponding pixel $p$. The final output operation is defined as 
\begin{equation}
    \bm{F'} = \sum_{i=0}^{H+W-1} \mathrm{\mathbf{A}} \cdot \mathrm{\mathbf{V}}_p + \bm{F}
\end{equation}
where $\bm{F'}\in \mathbb{R}^{C\times W \times H}$, and using residual operation to robust contextual information of spatial attention. \textcolor{deepred}{Sparse focus attention with fixed length follows the same process as described above}, but with a fixed length of row or column to join attention calculation.

Integrating sparse focus attention with other convolution operations into a Sparse Focus Transformer (SFT) network involves using sparse attention mechanisms within the transformer architecture. During the sparse encoder operation, feature maps $\bm{F_1}$ and $\bm{F_2}$, corresponding to different time steps, are passed through the Sparse Focus Transformer for feature localization and extraction. \textcolor{deepred}{The output results, $\bm{F_1}'$ and $\bm{F_2}'$, are then concatenated to form the final output $\bm{I_{1,2}}$.
The Sparse Focus Transformer operation can be expressed as follows:}
\textcolor{deepred}{
\begin{equation}
    \bm{F_1}' = \mathrm{SFT}(\bm{F_1})
\end{equation}
\begin{equation}
    \bm{F_2}' = \mathrm{SFT}(\bm{F_2})
\end{equation}
}
\textcolor{deepred}{where $\mathrm{SFT}(\cdot)$ represents the Sparse Focus Transformer operation applied to the input feature maps $\bm{F_1}$ and $\bm{F_2}$.}

\textcolor{deepred}{
After processing, the resulting feature maps are concatenated to form the final output:}
\textcolor{deepred}{
\begin{equation}
    \bm{I_{1,2}} = \mathrm{Concat}(\bm{F_1}', \bm{F_2}')
\end{equation}
}
\textcolor{deepred}{The concatenated output $\bm{I_{1,2}}$ combines the features processed by the Sparse Focus Transformer from different time steps.By these operations, the network progressively refines its focus on the most relevant features through multiple passes. This iterative process enhances the network's ability to extract and represent meaningful information from the input data, leading to improved task performance.}

\subsection{\textcolor{deepred}{Change Caption Generator}}
The techniques for generating descriptions are mainly divided into three categories: template-based approaches, retrieval-based approaches, and sequence generation-based approaches. To precisely describe and characterize the differences between image pairs, we endorse the sequence generation-based approaches. These approaches employ a transformer-based decoder, a method that has gained significant traction in numerous contemporary natural language processing tasks.

\begin{figure}[htb]
    \centering
    \includegraphics[width=\linewidth]{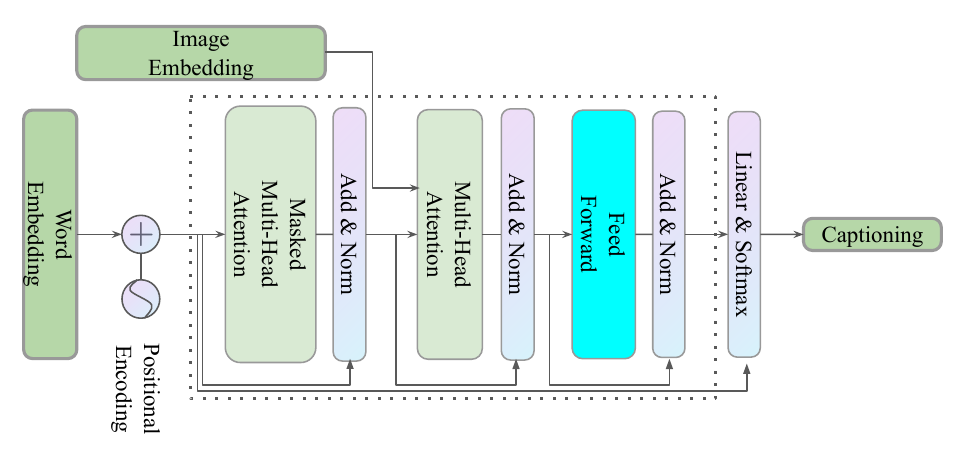}
    \caption{Visualization of caption generator}
    \label{fig7}
\end{figure}

The decoder consists of several layers of transformers, each featuring a masked multi-head attention sublayer and a feed-forward network. To ensure the continuity of information propagation and bolster the model's robustness, these sublayers are enhanced with residual connections and layer normalization techniques. Ultimately, the output embedding is produced through a linear layer, followed by the application of a softmax activation function. A visual depiction of this architecture is provided in Figure \ref{fig7}.

During the training phase, in order to prepare change descriptions for the caption decoder, the text tokens undergo an initial mapping process to transform them into word embeddings via an embedding layer $E_{embed}$. $E_{pos}$ represents the positional embedding computed using a sin/cos function with unique frequencies and phases assigned to each token's position\cite{devj}. $\bm{T}_{embed}$ refers to convert initial tokens $\bm{t}$ into word embeddings. The initial inputs of the transformer decoder can be acquired through the following procedure:
\begin{equation}
    \bm{T}_{embed} = E_{embed}(\bm{t}) + E_{pos}
\end{equation}
Then, we feed the aforementioned inputs into the masked multi-head attention layers. The computational process can be described as follows:
\begin{equation}
    \bm{Head}_l = \mathrm{Attention} (\bm{T}_{embed}^{i-1} \bm{W}_{l}^{Q}, \bm{T}_{embed}^{i-1}\bm{W}_{l}^{K}, \bm{T}_{embed}^{i-1}\bm{W}_{l}^{V})
\end{equation}
where 
\begin{equation}
    Attention(\bm{Q, K, V}) = Softmax(\frac{\bm{QK}^T}{\sqrt{d}})\bm{V}
\end{equation}
next is 
\begin{align}
    \bm{T}_{\text{img}} & = \mathrm{MHA}(\bm{T}_{\text{embed}}^{i-1}, \bm{T}_{\text{embed}}^{i-1}, \bm{T}_{\text{embed}}^{i-1}) \nonumber \\
                           & = \mathrm{Concat}(\bm{Head}_{1}, \dots, \bm{Head}_{h}) \cdot \bm{W}^{O}
\end{align}
in the masked multi-head attention sub-layer(MHA), $h$ represents the number of heads. The weight matrices $\bm{W}_{l}^{Q}, \bm{W}_{l}^{K}$ and $\bm{W}_{l}^{V}$ each of size $d_{embed}\times d_{embed/h}$ are trainable parameters associated with the $l$-th head. Here, $d_{embed}$ denotes the embedding dimension. $\bm{W}^{O} \in \mathbb{R}^{d_{embed}\times {d_{embed}}}$, a trainable weight matrix of size 
 , performs linear projection to adjust the output feature dimension. The ultimate output of the transformer decoder layer is attained by incorporating the input word embedding with the output of the feed-forward network via a residual connection:
\begin{equation}
     \bm{T}_{text} = \mathrm{FN}(\mathrm{Caption}_{text-img}) + \bm{T}_{embed}^{i-1}
\end{equation}
After the aforementioned operations, the word embedding is passed through a linear layer(LN) followed by a softmax activation function to derive the word probabilities, the final caption denoted as
\begin{equation*}
    \mathrm{Caption_{T}} = \mathrm{Softmax}(\mathrm{LN}(\bm{T}_{text}))
\end{equation*}
for $\mathrm{Caption_{T}=[\hat{\mathbf{t_1}},\hat{\mathbf{t_2}},...,\hat{\mathbf{t_n}}]}\in \mathbb{R}^{n\times m}$ where $n$ is the length of the generated caption and $m$ is the size of the vocabulary, $\hat{\mathbf{t_i}}$ is the predicted probability of generating the $i$-th word.

In the validation and testing stages, the SparseFocus Transformer (SFT) network employs an autoregressive method for generating captions from input image pairs. The decoding process begins with the ``START" token and utilizes the encoder's relevant image features to generate the next token. This is followed by the production of logits through a linear layer and the calculation of probabilities using softmax. The decoder integrates the encoder's output and previously generated tokens throughout this process to predict the next tokens until the ``END" token is reached.
\section{Experiments}
\label{sec4}
To perform a thorough evaluation of the proposed SFT approach, we conducted comparisons with existing remote sensing change captioning methods on two benchmark remote sensing change captioning (RSCC) datasets. Furthermore, we provide evidence to demonstrate that our proposed method not only enables rapid predictions with low complexity but also exhibits high accuracy.
\subsection{Datasets}
 \begin{itemize}
     \item LEVIR-CC Dataset: The LEVIR-CC dataset originates from a building change detection dataset consisting of 637 very high-resolution (0.5 m/pixel) bitemporal images sized $1,024 \times 1,024$ \textcolor{deepred}{\cite{liuc}}. To adapt it for use in RSCC, the LEVIR-CC dataset was curated by segmenting 10,077 small bitemporal tiles sized 256 × 256 pixels, with each tile annotated as containing changes or no changes. The dataset comprises 5,038 image pairs depicting changes and 5,039 pairs without changes, with each pair accompanied by five distinct sentence descriptions delineating the nature of changes between the two acquisitions. The maximum sentence length is 39 words, with an average of 7.99 words.
     \item Dubai-CC Dataset: The Dubai-CC dataset offers a detailed portrayal of urbanization changes within the Dubai region. To ensure precise identification and description of changes, the original images are partitioned into 500 tiles sized 50 $\times$ 50, with five change descriptions annotated for each small bitemporal tile, referencing Google Maps and publicly available documents. The dataset comprises 2,500 distinct descriptions, with a maximum length of 23 words and an average length of 7.35 words. Experimental configurations detailed in \cite{hoxha} were adopted, with the dataset divided into training, validation, and testing sets, containing 300, 50, and 150 bitemporal tiles, respectively.
 \end{itemize}
\subsection{Experimental Setup}
Evaluation Metrics: The efficacy of the captioning model hinges on its ability to produce descriptive sentences that align well with human judgments regarding differences between bitemporal images. To gauge this alignment, automatic evaluation metrics are employed to quantify the accuracy of the generated sentences against annotated reference sentences. In our study, we utilized four standard metrics prevalent in both image captioning \cite{stm}\cite{huangy} and change captioning \cite{jij}\cite{akk} domains:
\begin{enumerate}
    \item BLEU-N (N=1,2,3,4):\cite{bleu} measures the precision of n-gram overlap between the generated caption and the reference captions. It calculates the precision for each n-gram size (up to N) and combines them using a geometric mean, penalized by a brevity penalty to account for shorter generated captions.
    \begin{equation}
        \text{BLEU-N} = \text{BP} \times \exp\left(\sum_{n=1}^N \frac{1}{N} \log p_n\right)\\
    \end{equation}
    Where $p_n$ is the precision of n-grams, and BP is the brevity penalty.
    \item ROUGE-L:\cite{rouge} computes the longest common subsequence (LCS) between the generated caption and the reference captions. It normalizes this by the length of the longer of the two sequences, providing a measure of how well the generated caption captures the content of the reference captions.
    \begin{equation}
        \text{ROUGE-L} = \frac{\text{LCS}(g, r)}{\text{max}(\text{len}(g), \text{len}(r))} \\
    \end{equation}
    where $\text{LCS(g, r)}$ is the length of the longest common subsequence between the generated caption $(g) $ and the reference caption $(r)$.
    \item METEOR:\cite{meteor} calculates the harmonic mean of precision and recall, incorporating stemming and synonymy in its evaluation. It penalizes for fragmentation by rewarding contiguous matches and accounts for recall through the harmonic mean.
    \begin{equation}
        \text{METEOR} = (1 - \alpha) \cdot \text{P} \cdot \text{R} \cdot \frac{\text{P} + \beta \cdot \text{R}}{\text{P} + \text{R}} \\
    \end{equation}
    Where P is precision, R is recall, and $\alpha$ and $\beta$ are tunable parameters.
    \item CIDEr-D:\cite{cider} computes the consensus between the generated caption and the reference captions based on TF-IDF weighted n-grams. It emphasizes capturing diverse and descriptive phrases in the generated caption, giving higher scores for more informative and varied descriptions.
    \begin{equation}
        \text{CIDEr-D} = \text{TF-IDF}(\text{gen}, \text{ref}) \\
    \end{equation}
    Where TF-IDF is the term frequency-inverse document frequency weighted similarity between the generated caption and the reference captions.
\end{enumerate}
These metrics assess the consistency between predicted and reference sentences, with higher scores indicating a closer resemblance and thus higher accuracy in captioning.

Experimental Details: The deep learning methodologies expounded in this study are implemented within the PyTorch framework and executed on single NVIDIA A5000 GPU with 24G memory. Training and evaluation procedures adhere to meticulous parameters: employing the Adam optimizer \cite{adam} with an initial learning rate of 0.0001 and a weight decay of 0.5, The training regimen spans about 40 epochs, with a batch size set at 32 for enhanced computational efficiency. Following each epoch, the model undergoes rigorous evaluation against the validation set.

\subsection{Results Analysis}
This study presents a comprehensive evaluation of the proposed algorithm on two distinct datasets, LEVIR\_CC and Dubai\_CC.

\textbf{Accuracy Analysis.} we will delve into a comprehensive examination of the models' performance, \textcolor{deepred}{focusing on their
accuracy using different metrics} and how effectively their results in the given dataset.

On the LEVIR\_CC dataset, our Sparse Focus Transformer method demonstrates remarkable performance, outperforming several existing methods such as DUDA, MCCFormers-S, MCCFormers-D, RSICCformer, \textcolor{deepred}{ATTENTIVE-S, without positional embedding initialization, and ATTENTIVE, with positional embedding initialization}, as illustrated in Table \ref{tab1}. Specifically, it achieves a BLEU-4 score of 62.87\%, underscoring its effectiveness in accurately capturing and describing changes between bitemporal images. Furthermore, our approach surpasses the current state-of-the-art (SOTA) methods in various other evaluation metrics, highlighting its superiority in change captioning tasks. This fully validates our previous analysis and further demonstrates the effectiveness of the proposed algorithm.
\begin{table*}[tp]
  \caption{Comparison of Methods' Performance on Multiple Evaluation Metrics on LEVIR-CC Dataset}
  \centering
  \begin{threeparttable}
  \resizebox{0.8\textwidth}{!}{%
  \begin{tabular}{l|ccccccc}
    \toprule
    \multirow{2}{*}{\textbf{Method}} & \multicolumn{7}{c}{\textbf{Metrics}} \\
    \cmidrule(lr){2-8}
    & \textbf{BLEU-1} & \textbf{BLEU-2} & \textbf{BLEU-3} & \textbf{BLEU-4} 
    & \textbf{METEOR} & \textbf{ROUGE-L} & \textbf{CIDEr-D} \\
    \midrule
    DUDA & 81.44 & 72.22 & 64.24 & 57.79 & 37.15 & 71.04 & 124.32 \\
    CC+RNN & - & - & - & - & - & - & - \\
    CC+SVM & - & - & - & - & - & - & - \\
    MCCFormers-S & 82.16 & 72.95 & 65.42 & 59.41 & 38.26 & 72.10 & 128.34 \\
    MCCFormers-D & 80.49 & 71.11 & 63.52 & 57.34 & 38.23 & 71.40 & 126.85 \\
    RSICCformer & 84.11 & 75.40 & 68.01 & 61.93 & 38.79 & 73.02 & 131.40 \\
    ATTENTIVE-S & 82.41 & 73.10 & 65.29 & 59.02 & 38.71 & 72.47 & 130.88 \\
    ATTENTIVE & \textbf{85.14} & \textbf{76.91} & \textbf{69.86} & \textbf{64.09} & 39.83 & 74.62 & 135.41 \\
    \midrule
    SparseFocus Full\tnote{*} & \textcolor{blue}{84.56} & \textcolor{blue}{75.87} & \textcolor{blue}{68.64} & \textcolor{blue}{62.87} & \textbf{39.93} & \textbf{74.69} & \textbf{137.05}\\
    \bottomrule
  \end{tabular}%
  }
  \begin{minipage}{0.8\textwidth}
    \justifying
    \footnotesize
    \tnote{*} \textcolor{deepred}{Note: It refers to the SparseFocus Full (R=1) model, which is the best model. R indicates the number of stacked Sparse Focus attention layers. A detailed analysis of the ablation experiments will be discussed later.}
  \end{minipage}
  \end{threeparttable}
  \label{tab1}
\end{table*}
\begin{table*}[tp]
  \caption{Comparison of Methods' Performance on Multiple Evaluation Metrics on DUBAI-CC Dataset}
  \centering
  \begin{threeparttable}
  \resizebox{0.8\textwidth}{!}{%
  \begin{tabular}{l|ccccccc}
    \toprule
    \multirow{2}{*}{\textbf{Method}} & \multicolumn{7}{c}{\textbf{Metrics}} \\
    \cmidrule(lr){2-8}
    & \textbf{BLEU-1} & \textbf{BLEU-2} & \textbf{BLEU-3} & \textbf{BLEU-4} 
    & \textbf{METEOR} & \textbf{ROUGE-L} & \textbf{CIDEr-D} \\
    \midrule
    DUDA & 58.82 & 43.59 & 33.63 & 25.39 & 22.05 & 48.34 & 62.78 \\
    CC-RNN-sub & 67.19 & 52.23 & 40.00 & 28.54 & 25.51 & 51.78 & 69.73 \\
    CC+SVM-sub & 70.71 & 57.58 & 46.10 & 35.50 & 27.60 & 56.61 & 82.96 \\
    MCCFormers-S & 52.97 & 37.02 & 27.62 & 22.57 & 18.64 & 43.29 & 53.81 \\
    MCCFormers-D & 64.65 & 50.45 & 39.36 & 29.48 & 25.09 & 51.27 & 66.51 \\
    RSICCformer & 67.92 & 53.61 & 41.37 & 31.28 & 25.41 & 51.96 & 66.54 \\
    ATTENTIVE-S & 65.75 & 51.87 & 42.01 & 33.09 & 25.17 & 52.06 & 72.23 \\
    ATTENTIVE & \textbf{72.04} & \textbf{60.18} & \textbf{50.84} & \textbf{41.70} & \textbf{28.92} & \textbf{58.66} & \textbf{92.49} \\
    \midrule
    SparseFocus Full \tnote{*} & \textcolor{blue}{67.30} & \textcolor{blue}{55.97} & \textcolor{blue}{47.00} & \textcolor{blue}{37.30} & \textcolor{blue}{26.32} & \textcolor{blue}{56.38} & \textcolor{blue}{91.59} \\
    \bottomrule
  \end{tabular}%
  }
  \begin{minipage}{0.8\textwidth}
    \justifying
    \footnotesize
    \tnote{*} \textcolor{deepred}{Note: It refers to the SparseFocus Full (R=1) model, which is the best model. R indicates the number of stacked Sparse Focus attention layers. A detailed analysis of the ablation experiments will be discussed later.}
  \end{minipage}
  \end{threeparttable}
  \label{tab2}
\end{table*}

In the comparison table presented in Table \ref{tab2}, our Sparse Focus Transformer method is evaluated against several other methods on the DUBAI-CC Dataset. \textcolor{deepred}{While our method performs reasonably well, it ranks second in overall performance metrics. The leading method, ATTENTIVE, achieves the highest scores across all evaluation metrics, indicating a stronger ability to capture the nuances of change between bitemporal images. Nevertheless, our Sparse Focus method demonstrates reasonably competitive performance, closely trailing the leading approach. Specifically, Sparse Focus achieves moderate scores in various metrics, such as BLEU-1, BLEU-2, ROUGE-L, and CIDEr-D, suggesting its potential in change captioning tasks.} Despite securing the second position, our method's performance underscores its efficacy and potential as a robust solution for change detection and description in remote sensing images.

\textbf{Parameter and Complexity Analysis.}
This section provides an in-depth analysis of model parameters and computational complexity, elucidating the intricate balance between model size, computational efficiency. By examining these factors, we gain valuable insights into the efficiency and effectiveness of the proposed approach.

\textcolor{deepred}{\textbf{Parameter \textcolor{deepred2}{and} Inference Analysis.}} The table in Table \ref{tab3} compares different methods based on their total parameters, encoder module parameters, and inference time for images sized 256 × 256 pixels. SparseFocus is prominently featured in this comparison.

While SparseFocus ranks second in terms of accuracy, it stands out as the top performer in terms of parameter efficiency and computational complexity. For example, the best model from our work, Sparse Focus Full (depicted by the grey line), has only about half the total parameters compared to the highest-accuracy model, ATTENTIVE. This advantage is especially pronounced in the image encoder module, where our encoder has over 90\% fewer parameters compared to the 648M parameters of the ATTENTIVE model's encoder. This considerable reduction in model size underscores its efficiency in terms of both memory occupation and resource utilization, positioning it as the most lightweight method among the approaches considered, especially given comparable accuracy.

Moreover, SparseFocus exhibits competitive performance in inference time, demonstrating its potential for fast and efficient inference. Although its inference time is marginally longer than the baseline, it still falls within a moderate range, ensuring that it is practical for real-world applications.

In summary, SparseFocus emerges as a promising solution for change detection tasks, excelling in parameter efficiency and computational complexity while maintaining competitive accuracy. Its lightweight nature and reasonable inference time make it a suitable option for applications with strict resource constraints or real-time processing requirements. This combination of efficiency and practicality underscores its potential for broader adoption in environments where hardware limitations and processing speed are critical considerations.

\begin{table*}[thb]
    \centering
    \caption{Comparison of Parameters and Inference Time with image size $256\times 256$}
    \label{tab3}
    \resizebox{0.8\textwidth}{!}{%
        \begin{tabular}{l|cccc}
            \toprule
            \textbf{Method} & \textbf{Total Parameters (M)} & \textbf{Encoder Module Parameters (M)} & \textbf{Inference Time (ms)} \\
            \midrule
            DUDA             & 140.35    & 1.15   & 30.18 \\
            CC-RNN-sub       & -         & -      & - \\
            CC-SVM-sub       & -         & -      & - \\
            MCCFormers-S     & 131.50    & 26.08  & 36.11 \\
            MCCFormers-D     & 131.43    & 26.08  & 35.71 \\
            RSICCformer      & 790.00    & 455.67 & 44.30 \\
            ATTENTIVE        & 1300.00   & 648.56 & 49.10 \\
            \midrule
            SparseFocus Fixed(R=1) & \textcolor{blue}{617.02}  & \textcolor{blue}{10.16} & \textcolor{blue}{25.01} \\
            SparseFocus Fixed(R=2) & \textcolor{blue}{628.05}  & \textcolor{blue}{21.19} & \textcolor{blue}{28.13} \\
            \midrule
            \rowcolor{gray!30}
            SparseFocus Full(R=1) & \textcolor{blue}{647.00}  & \textcolor{blue}{40.14} & \textcolor{blue}{31.02} \\
            SparseFocus Full(R=2) & \textcolor{blue}{687.10}  & \textcolor{blue}{80.18} & \textcolor{blue}{38.80} \\
            \bottomrule
        \end{tabular}%
    }
\end{table*}
\begin{table*}[htbp]
    \centering
    \caption{Comparison of Model Memory Usage and Computational Efficiency}
    \label{tab4}
    \resizebox{0.8\textwidth}{!}{%
    \begin{tabular}{l|ccc}
        \toprule
        \textbf{Model} & \textbf{Training GPU Memory (GB)} & \textbf{Inference GPU Memory (GB)} & \textbf{MACs (GB)} \\
        \midrule
        RSICCformer & 2.57 & 1.10 & 150.01 \\
        ATTENTIVE & 12.35 & 2.53 & 464.29 \\
        \midrule
        SparseFocus Fixed(R=1) & 8.52 & 0.60 & 11.30 \\
        SparseFocus Fixed(R=2) & 8.77 & 0.61 & 19.10 \\
        \midrule
        \rowcolor{gray!30}
        SparseFocus Full(R=1) & 9.43 & 0.66 & 21.47 \\
        SparseFocus Full(R=2) & 9.47 & 0.70 & 42.95 \\
        \bottomrule
    \end{tabular}
    }
\end{table*}
\textcolor{deepred}{\textbf{GPU Memory and MACs Analysis.}} Table \ref{tab4} provides a comparative analysis of various models, focusing on their memory usage and computational efficiency, specifically regarding training GPU memory, inference GPU memory, and Multiply-Accumulate operations (MACs) for images sized 256 × 256 pixels.

SparseFocus exhibits notable advantages in terms of GPU memory utilization for both training and inference. For example, the best model from our study (the grey line) requires approximately 9GB for training, whereas the ATTENTIVE model requires more than 12GB. Additionally, the inference GPU memory usage for SparseFocus is around 0.6GB, significantly lower than other models, which can exceed 1GB or even 2GB. This considerable reduction in memory usage demonstrates the efficiency of SparseFocus in leveraging computational resources during both model training and inference.

Furthermore, SparseFocus presents significantly lower Multiply-Accumulate operations (MACs), indicating superior computational efficiency during model execution. Its MACs count is only 21.47GB, far less than the RSICCformer and ATTENTIVE models, which require 150GB and 464GB respectively. This reduction in computational overhead indicates that SparseFocus achieves comparable or even better performance with substantially lower resource requirements.

Overall, these results suggest that SparseFocus is not only efficient in terms of memory usage but also demonstrates superior computational efficiency. This makes it a highly attractive option for applications where resources are limited, or where computational efficiency is a critical factor. By maintaining high performance with reduced computational overhead, SparseFocus emerges as a compelling choice for a wide range of scenarios, particularly in environments with strict resource constraints or where speed and efficiency are paramount.

\subsection{Ablation Studies}
\textbf{LEVIR\_CC Ablation.} In our study, we implemented the SparseFocus mechanism with attention kernel length $l$ determined as $l=w$, where $w = h = 8$ is the size of our feature maps $F$. This led to a full attention length for \textbf{Sparse Focus Full}, for the fixed length $1/2 w$ is named \textbf{Sparse Focus Fixed} ensuring efficient information capture while minimizing computational complexity.

In our subsequent experiments, we evaluated the performance of the proposed methods, Sparse Focus Full and Sparse Focus Fixed, across LEVIR\_CC datasets. In this context, $R$ represents the number of stacked attention layers. Table \ref{tab5} provides a comprehensive summary of the results obtained from these experiments, demonstrating the relative efficacy of these two approaches.
One of the most striking findings was that the Sparse Focus Full approach, with a single stack ($R = 1$) of full-length attention, achieved the highest scores among the tested configurations. Specifically, this approach attained a BLEU-4 score of 62.87 and a CIDEr-D score of 137.05, indicating a robust capability for capturing semantic changes and outperforming alternative methods in key evaluation metrics.
Interestingly, these results also highlighted a potential trade-off associated with stacking multiple layers of SparseFocus. While increased depth could theoretically enhance the model's learning capacity, our findings suggest that this approach might lead to diminishing returns due to the introduction of additional sparsity. In particular, the data indicated that increased sparsity could result in a loss of effective information, especially when dealing with images of dimensions $256 \times 256$. This observation motivated the adoption of a short attention length in Sparse Focus Fixed is only $l = 4$, without shorter length, aiming to maintain a balance between model complexity and information retention.

In summary, the results from these experiments emphasize the importance of carefully considering model architecture, especially when utilizing attention mechanisms. The success of Sparse Focus Full with a single stack suggests that simpler configurations can often yield superior outcomes, while excessive complexity may adversely impact performance. These insights will guide future research into optimal configurations for attention-based models, ensuring a balance between precision and efficiency.

\begin{table*}[htbp]
    \centering
    \caption{Performance Evaluation and Change Accuracy of SparseFocus Methods on LEVIR-CC dataset}
    \label{tab5}
    \resizebox{\textwidth}{!}{%
    \begin{tabular}{c|ccccccc|ccc}
        \toprule
        \multirow{3}[0]{*}{\textbf{Method}} & \multicolumn{7}{c}{\textbf{Captioning Metrics}} & \multicolumn{3}{c}{\textbf{Image Metrics}} \\
        \cmidrule(lr){2-8} \cmidrule(lr){9-11}
        & \textbf{BLEU-1} & \textbf{BLEU-2} & \textbf{BLEU-3} & \textbf{BLEU-4} & \textbf{METEOR} & \textbf{ROUGE-L} & \textbf{CIDEr-D} & \textbf{Change Accuracy} & \textbf{No-change Accuracy} & \textbf{Total Accuracy} \\
        \midrule
        SparseFocus Fixed(R=1) & 82.56 & 72.87 & 65.64 & 60.87 & 36.93 & 71.69 & 134.05 & 85.36 & 95.73 & 90.55 \\
        SparseFocus Fixed(R=2) & 80.56 & 70.87 & 63.64 & 58.87 & 34.93 & 69.69 & 132.05 & 82.66 & 93.43 & 88.05 \\
        \rowcolor{gray!30}
        SparseFocus Full(R=1) & 84.56 & 75.87 & 68.64 & \textbf{62.87} & \textbf{39.93} & \textbf{74.69} & \textbf{137.05} & 87.86 & 98.03 & 92.95 \\
        SparseFocus Full(R=2) & \textbf{85.00} & \textbf{76.14} & \textbf{68.64} & 62.52 & 39.46 & 73.88 & 134.18 & 90.66 & 95.65 & 93.16 \\
        \bottomrule
    \end{tabular}%
    }
\end{table*}
\begin{table*}[htbp]
    \centering
    \caption{Performance Evaluation and Change Accuracy of SparseFocus Methods on DUBAI-CC dataset}
    \label{tab6}
    \resizebox{\textwidth}{!}{%
    \begin{tabular}{c|ccccccc|ccccc}
        \toprule
        \multirow{3}[0]{*}{\textbf{Method}} & \multicolumn{7}{c}{\textbf{Captioning Metrics}} & \multicolumn{3}{c}{\textbf{Image Metrics}} \\
        \cmidrule(lr){2-8} \cmidrule(lr){9-11}
        & \textbf{BLEU-1} & \textbf{BLEU-2} & \textbf{BLEU-3} & \textbf{BLEU-4} & \textbf{METEOR} & \textbf{ROUGE-L} & \textbf{CIDEr-D} & \textbf{Change Accuracy} & \textbf{No-Change Accuracy} & \textbf{Total Accuracy} \\
        \midrule
        \rowcolor{gray!30}
        SparseFocus Full(R=1) & 67.30 & \textbf{55.97} & \textbf{47.00} & \textbf{37.30} & 26.32 & \textbf{56.38} & \textbf{91.59} & 72.17 & 74.29 & 73.23 \\
        SparseFocus Full(R=2) & \textbf{68.28} & 53.61 & 43.96 & 36.28 & \textbf{26.40} & 54.72 & 86.25 & 76.52 & 77.14 & 76.83 \\
        \bottomrule
    \end{tabular}%
    }
\end{table*}
\begin{table*}[thb]\centering
    \caption{Ablation Study on Model Components and Performance Evaluation on LEVIR-CC dataset}
    \label{tab7}
    \begin{threeparttable}
    \resizebox{0.8\textwidth}{!}{
    \large
    \begin{tabular}{c|ccc|ccccccc}
        \toprule
       \textbf{ID} & \textbf{Extractor} & \textbf{Image Encoder} & \textbf{Caption Decoder} & \textbf{BLEU-1} & \textbf{BLEU-2} & \textbf{BLEU-3} & \textbf{BLEU-4} & \textbf{Meteor} & \textbf{Rouge} & \textbf{Cider-D} \\
        \midrule
        1 & ResNet50 & \makecell[c]{SFT\\(DEFAULT)} & \makecell[c]{ChangeCaptionGenerator\\(DEFAULT)} & 83.35 & 75.15 & 67.66 & 61.53 & 37.75 & 71.93 & 124.89 \\
        \hline
        2 & VGG16\_BN & \makecell[c]{SFT\\(DEFAULT)} & \makecell[c]{ChangeCaptionGenerator\\(DEFAULT)} & 80.30 & 70.64 & 62.84 & 56.61 & 37.59 & 71.04 & 124.84 \\
        \hline
        3 & VIT16 & \makecell[c]{SFT\\(DEFAULT)} & \makecell[c]{ChangeCaptionGenerator\\(DEFAULT)} & 79.49 & 70.56 & 63.68 & 58.37 & 36.34 & 70.37 & 122.83 \\
        \hline
        4 & \makecell[c]{ResNet101\\(DEFAULT)} & VIT16 & \makecell[c]{ChangeCaptionGenerator\\(DEFAULT)} & 83.57 & 75.02 & 67.86 & 62.07 & 39.09 & 73.36 & 132.56 \\
        \hline
        5 & \makecell[c]{ResNet101\\(DEFAULT)} & \makecell[c]{SFT\\(DEFAULT)} & SFT & 77.98 & 68.33 & 60.18 & 54.28 & 35.06 & 
        68.78 & 122.62 \\
        \hline
        Ours & \makecell[c]{ResNet101\\(DEFAULT)} & \makecell[c]{SFT\\(DEFAULT)} & \makecell[c]{ChangeCaptionGenerator\\(DEFAULT)} & \textbf{84.56} & \textbf{75.87} & \textbf{68.64} & \textbf{62.87} & \textbf{39.93} & \textbf{74.69} & \textbf{137.05}\\
        \bottomrule
    \end{tabular}
    }

  \end{threeparttable}
\end{table*}

The table \ref{tab5} illustrates the change accuracy of Sparse Focus methods on the LEVIR-CC dataset. Notably, Sparse Focus Full  consistently outperforms Sparse Focus Fixed, emphasizing the effectiveness of utilizing the full-length attention mechanism. Specifically, Sparse Focus Full achieves superior change accuracy and total accuracy compared to Sparse Focus Fixed. Overall, Sparse Focus Full demonstrates superior performance in accurately detecting changes while maintaining high classification accuracy. 

\textbf{DUBAI\_CC Ablation.} The performance evaluation on the DUBAI-CC dataset highlights the effectiveness of Sparse Focus methods across various evaluation metrics, as depicted in Table \ref{tab6}. Specifically, Sparse Focus Full with $R=1$ stack layers outperforms $R=2$ in terms of BLEU scores, METEOR, ROUGE-L, and CIDEr-D. Notably, Sparse FocusFull achieves the highest BLEU-4 score of 37.30 and the highest CIDEr-D score of 91.59, indicating its robustness in capturing semantic changes and generating descriptive captions. 
It's worth noting that due to the small size of the dataset (50$\times$50), we upsampled the images to 256$\times$256 to ensure consistency in size and better processing of image features, which could have contributed to the improved performance of SparseFocus. This preprocessing step facilitates a more comprehensive analysis of the model's capabilities in handling diverse image characteristics and extracting meaningful information. Overall, these findings underscore the efficacy of Sparse Focus in change captioning tasks, particularly when equipped with a full-length attention mechanism.

Based on the results presented in Table \ref{tab6}, Sparse Focus methods demonstrate superior performance in change accuracy on the dubai-CC dataset. Specifically, Sparse Focus Full($R=1$) consistently outperforms SparseFocus Full($R=2$) across all evaluated metrics, achieving higher accuracy in both change and no-change detection tasks. 
The substantial improvement observed with full receptive fields suggests the significance of incorporating complete contextual information for accurate change detection. This finding underscores the effectiveness of SparseFocus in leveraging comprehensive spatial dependencies within the image, enabling more precise identification of changes between bitemporal images.
\begin{figure*}[htp]
    \includegraphics[width=0.5\textwidth]{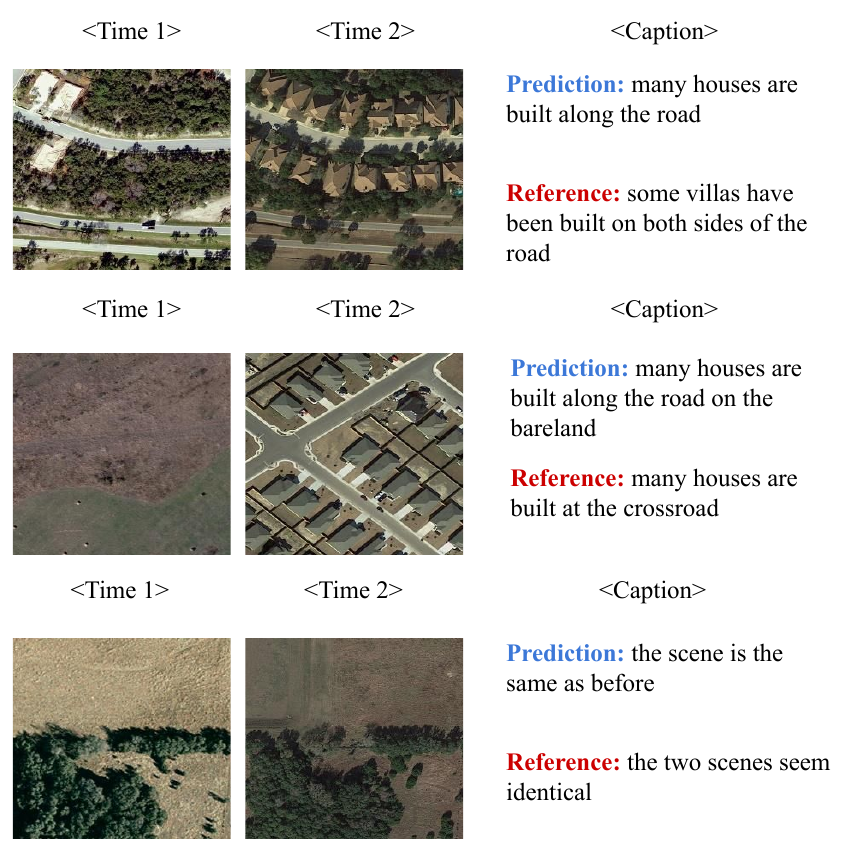}
    \includegraphics[width=0.5\textwidth]{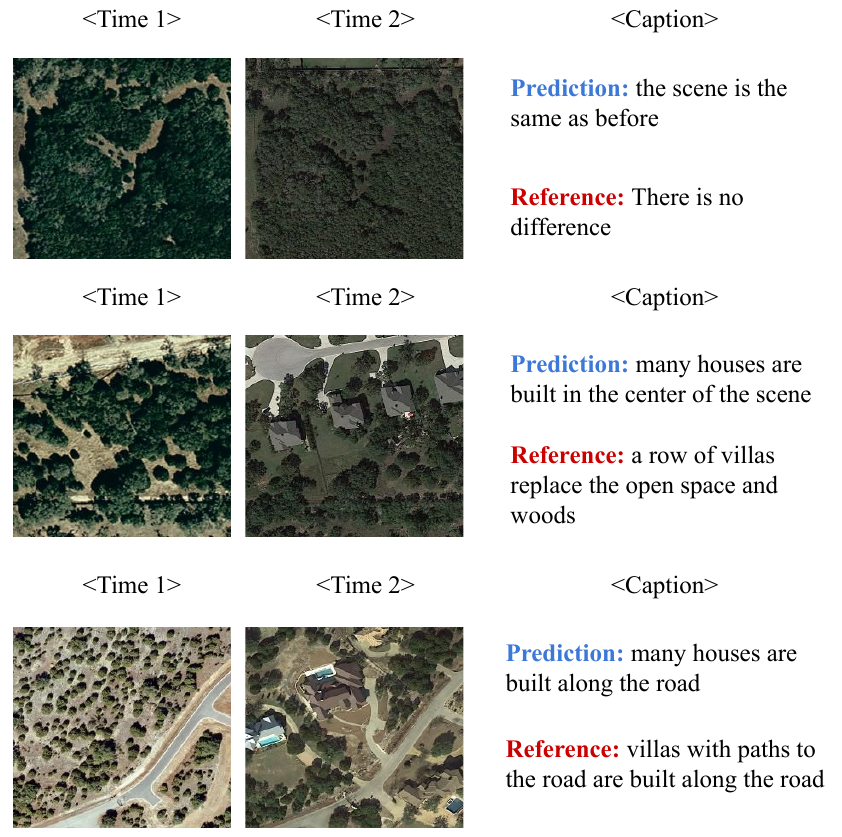}
    \caption{Visualized Image and Captioning examples generated by SFT on the LEVIR-CC dataset}
    \label{fig8}
\end{figure*}
\begin{figure*}[htp]
    \includegraphics[width=0.5\textwidth]{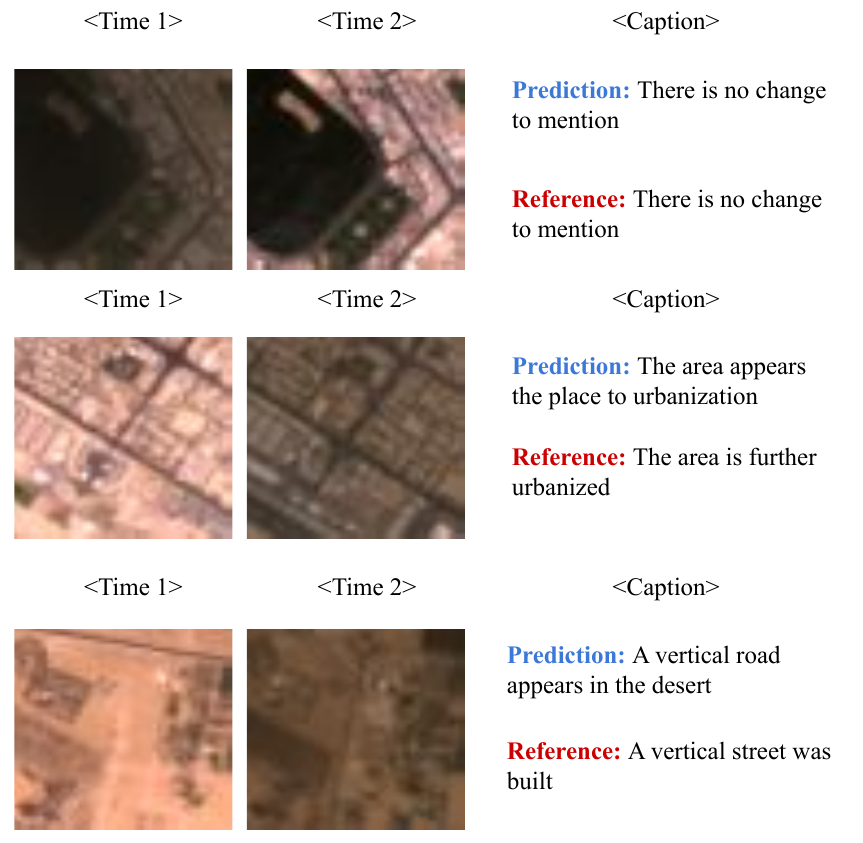}
    \includegraphics[width=0.5\textwidth]{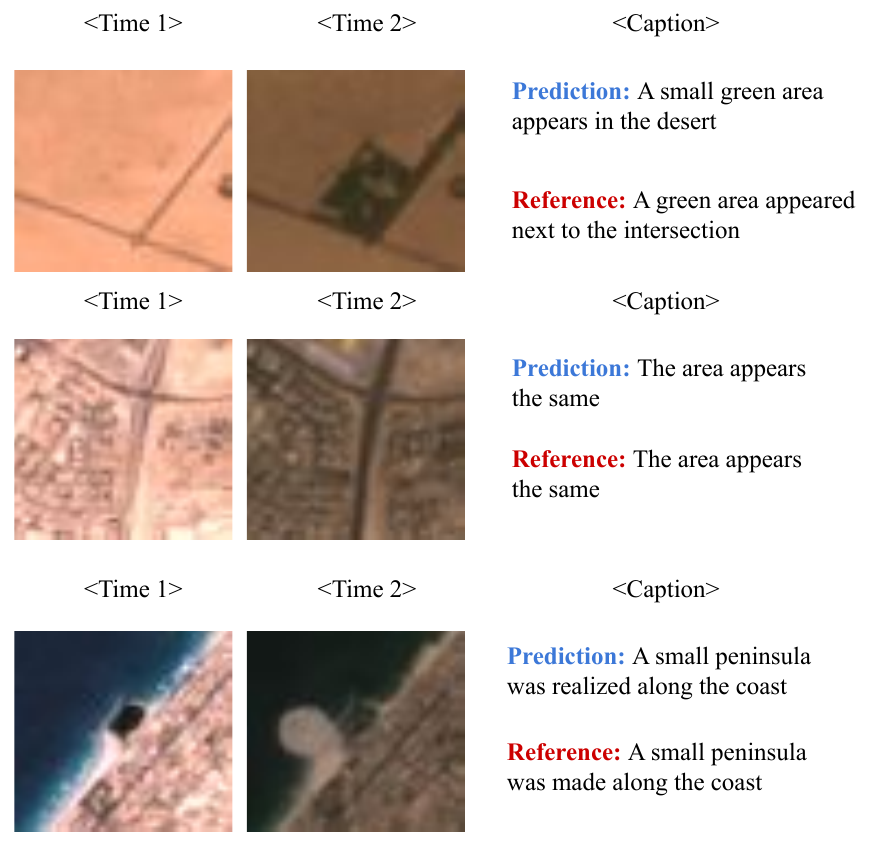}
    \caption{Visualized Image and Captioning examples generated by SFT on the DUBAI-CC dataset}
    \label{fig9}
\end{figure*}

However, Tables~\ref{tab5} and \ref{tab6} also reveal that in the Sparse Focus Full series, stacking a single attention layer yields better accuracy in the final captioning result, whereas stacking two layers results in higher performance in image transformation detection. This observation suggests that, despite a consistent decoder, the optimal output of the image encoder does not necessarily contribute to a more accurate captioning by the decoder. 
There might be several reasons for this inconsistency. Firstly, conventional similarity computation methods, such as cosine similarity, might not effectively integrate the image and text information accurately. This limitation could hinder the decoder's ability to generate captions that accurately reflect the content of the image. Secondly, the text decoder might require further design considerations to effectively decode the image information output by the image encoder, indicating that a more refined approach in integrating the encoder's output with the decoder's architecture could lead to improved captioning accuracy.
Overall, this analysis underscores the need to re-examine the relationship between image encoding and text decoding to ensure that the information flow between these two components is optimized. By addressing these challenges, future work can focus on developing more robust methodologies to achieve a seamless fusion of image and text information, thereby enhancing the accuracy of captioning.

\textcolor{deepred}{\textbf{Model Architecture Ablation:} We conducted ablation experiments by employing different networks for the Extractor, Image Encoder, and Caption Decoder components of the model to further validate the effectiveness of the method proposed in this paper. As shown in the Table~\ref{tab7}, when different backbone networks were utilized as the Extractor for feature extraction, a decrease in accuracy was observed. Furthermore, we employed the `vit\_base\_patch16\_224' model as the Image Encoder and made slight adjustments to the ViT model to better align it with our method. As indicated in the fourth row of the table, the accuracy of the experimental results is second only to our proposed method. However, the model's parameter count reached approximately 1.4G, with training GPU memory exceeding 15GB. Additionally, we integrated the Sparse Focus module into the caption generator. The results, as seen in the fourth row of the table, reveal that the accuracy was the lowest, likely due to the sequential regression and prediction of each word during caption generation. If the final decoder is too sparse, it may struggle to accurately predict the next word, thus negatively impacting overall experimental accuracy.}

\subsection{Qualitative Visualization}
To assess the effectiveness of the change captions produced by our proposed SparseFocus Transformer (SFT) method, we performed a qualitative evaluation by selecting representative scenes from both the Dubai-CC and LEVIR-CC datasets. These scenes are depicted in Figures \ref{fig8} and \ref{fig9}. In each figure, TIME1 and TIME2 correspond to remote sensing images captured at different times in the same area. The captions on the right represent the predictions generated by our model, while the captions below labelled as 'reference' serve as the ground truth labels.

\subsection{Limitation \& Future work}
Through the research presented in this paper, regarding the current issues and future research directions in the captioning of change detection for remote sensing images, the main focus should be the following aspects. First, we address the problem of multimodal fusion to more accurately and effectively extract representations of individual modalities. Only when modalities are sufficiently clear and accurate can they be effectively fused and aligned. Second, for the remote sensing image change captioning task (RSICC), exploring novel solution architectures instead of the existing three-stage paradigm. Third, exploring additional multimodal fusion techniques, such as similarity calculation methods, will facilitate accurate matching between image and text modalities.

\section{Conclusion}
\label{sec5}
In summary, our Sparse Focus Transformer (SFT) network offers a compelling solution to the task of change captioning in remote sensing imagery. Through rigorous experimentation, we demonstrate its effectiveness in accurately describing changes while significantly reducing computational complexity and parameter count compared to existing methods. Our approach not only advances change captioning technology but also contributes to efficient multimodal modeling, with potential applications in various industries and domains requiring high-dimensional data interaction.


%





\ifCLASSOPTIONcaptionsoff
  \newpage
\fi



\bibliographystyle{IEEEtran}
\bibliography{IEEEabrv,main}
%

%

\end{document}